\documentclass[10pt,twocolumn,letterpaper]{article}

\usepackage[pagenumbers]{wacv}      

\usepackage{graphicx}
\usepackage{amsmath}
\usepackage{amssymb}
\usepackage{booktabs}
\usepackage{enumitem}
\usepackage{svg}
\usepackage{makecell}
\usepackage{multirow}
\usepackage{colortbl}
\usepackage{pifont}

\usepackage[pagebackref,breaklinks,colorlinks]{hyperref}

\newcommand{\shortnamenospace}{HFGaussian}
\newcommand{\shortname}{HFGaussian }
\newcommand{\longname}{HFGaussian: Learning Generalizable Gaussian Human with Integrated Human Features}
\usepackage[capitalize]{cleveref}
\crefname{section}{Sec.}{Secs.}
\Crefname{section}{Section}{Sections}
\Crefname{table}{Table}{Tables}
\crefname{table}{Tab.}{Tabs.}

\begin{document}

\title{\longname}

\author{Arnab Dey$^1$\thanks{Contact email: \href{mailto:adey@i3s.unice.fr}{adey@i3s.unice.fr}} \and Cheng-You Lu$^2$ \and Andrew I. Comport$^1$  \and Srinath Sridhar$^3$ \and Chin-Teng Lin$^2$ \and Jean Martinet$^1$\and \\
$^1$I3S-CNRS/Universit\'e C\^ote d'Azur \quad $^2$University of Technology Sydney \quad $^3$Brown University
}

\maketitle

\begin{abstract}
   Recent advancements in radiance field rendering show promising results in 3D scene representation, where Gaussian splatting-based techniques emerge as state-of-the-art due to their quality and efficiency. 
   Gaussian splatting is widely used for various applications, including 3D human representation.
   However, previous 3D Gaussian splatting methods either use parametric body models as additional information or fail to provide any underlying structure, like human biomechanical features, which are essential for different applications. 
   In this paper, we present a novel approach called HFGaussian that can estimate novel views and human features, such as the 3D skeleton, 3D key points, and dense pose, from sparse input images in real time at 25 FPS. 
   The proposed method leverages generalizable Gaussian splatting technique to represent the human subject and its associated features, enabling efficient and generalizable reconstruction. 
   By incorporating a pose regression network and the feature splatting technique with Gaussian splatting, HFGaussian demonstrates improved capabilities over existing 3D human methods, showcasing the potential of 3D human representations with integrated biomechanics. 
   We thoroughly evaluate our HFGaussian method against the latest state-of-the-art techniques in human Gaussian splatting and pose estimation, demonstrating its real-time, state-of-the-art performance. 
\end{abstract}
\vspace{-15pt}
\section{Introduction}\label{sec:intro}
Generating virtual photorealistic 3D human avatars is a long-standing challenge in the field of computer vision~\cite{aitpayev2012creation, ichim2015dynamic}. 
These 3D models have diverse applications in fields such as augmented reality, virtual reality~\cite{li20193d}, entertainment, and the medical domain~\cite{singh2017darwin, cakmak2020human}. 
The task of reconstructing a complete 3D human model with integrated structural properties~\cite{buades2004human, zuo2020sparsefusion} in real time from images alone presents significant challenges. 
Classical approaches rely on complex multiview capture systems and body markers~\cite{Dyna:SIGGRAPH:2015, h36m_pami} to obtain 3D models of humans, incorporating structural properties such as 3D pose involve fitting parametric body models such as SMPL~\cite{SMPL:2015} and STAR~\cite{osman2020star}. 
However, these methods require substantial resources and computational effort to generate each 3D model. 

In recent years, radiance field rendering becomes significantly popular~\cite{tewari2022advances,wu2024recent} for the scene representation capabilities. 
More recently, 3D Gaussian splatting (3DGS)~\cite{kerbl20233d} provides a new research direction and demonstrates notable improvements compared to neural rendering-based methods.
3D Gaussian splatting proposes a novel explicit representation that represents the scene using a set of 3D Gaussians for point-based rendering. 
The efficient representation of Gaussian splatting makes it particularly well-suited for real-time rendering applications. 
Subsequent researches apply 3DGS to various applications~\cite{matsuki2024gaussian, chen2024text, wu20244d} including 3D human reconstructions~\cite{zheng2023gps, hu2023gauhuman}. 
The existing methods employing 3D Gaussian Splatting (3DGS) for human avatar reconstruction either rely on parametric body models or fail to incorporate any underlying biomechanical features crucial for downstream applications~\cite{yang2021s3, hohne20123d}. 

In this work, we propose a novel, generalizable approach for estimating a 3D human representation with integrated 3D pose and dense pose in real time, given sparse input images of the human subject. 
The proposed method, named Human Feature Gaussian (HFGaussian), uses Gaussian splatting to represent the human subject and its associated biomechanical features\footnote{In this study, "biomechanical features" refer to components of the human musculoskeletal system, such as bones, muscles, ligaments, and joint locations, which are critical for human movement and function.}, which include the 3D skeleton, 3D keypoints, and dense pose. 
These biomechanical properties are essential for recreating natural human movements and interactions in the virtual world~\cite{fortini2023markerless, jiang2019synthesis}. 
One straightforward approach to representing human features while maintaining real-time rendering speed, is to directly parameterize the 3D Gaussian with additional human features.
However, we point out that simply parameterizing the 3D Gaussian with these human features results in sub-optimal performance, as the same parameters like opacity, scaling, and rotation factor are not suitable for different human features.
Instead of directly parameterizing the 3D Gaussian with human features, inspired by feature splatting~\cite{martins2024feature}, we learn these human features by optimizing additional feature parameters for each 3D Gaussian, which are then decoded into human features after rendering.

Regarding 3D pose estimation, we find that even a subset of 3D Gaussians can serve as an effective point cloud for 3D pose estimation using a novel pose regression network based on DGCNN~\cite{wang2019dynamic} and PointNet~\cite{qi2017pointnet}.

In conclusion, we propose HFGaussian, a human-centric Gaussian framework that enables real-time representation of human features through 3D Gaussians.  
Using GPS-Gaussian~\cite{zheng2023gps} as the backbone, HFGaussian can generalize to unseen human data without any fine-tuning. 
Building on this foundation, we additionally introduce feature splatting~\cite{martins2024feature} to overcome the performance constraints of using the same set of Gaussians for various human features which may have different frequencies. 
Furthermore, HFGaussian employs a novel pose regression network to estimate the 3D pose from a subset of the 3D Gaussians, ensuring efficient estimation. 
HFGaussian is capable of simultaneously rendering novel poses, corresponding 3D poses, and human features in real time. 
Although this study focuses on human pose estimation, we believe that the HFGaussian can be extended to include other human features, such as body part segmentation.

To evaluate our proposed method, we train our method in a large amount of human data generated from human scans and evaluate in real-world data. 
To the best of our knowledge, this is the first method to estimate 3D humans with biomechanics features and 3D pose in real time directly from images.  
The contributions of this work can be summarized as follows:
\begin{itemize}[noitemsep,topsep=0pt]
    \item[$\bullet$] We present a novel generalizable approach named HFGaussian that is capable of estimating human features and 3D human pose.
    \item[$\bullet$] The proposed method has demonstrated its ability to estimate 3D pose using a novel pose regression network and human features using feature splatting.
    \item[$\bullet$] We propose a generalizable approach to predicting human features, 3D pose, photometric, and geometric representations from 2D sparse images in real time.
    \item[$\bullet$] Our extensive experimental analysis across 3 datasets validates the applicability and versatility of our method. 
    
\end{itemize}
\section{Related works}
The proposed method \shortname uses the Gaussian splatting technique to estimate 3D human avatars with integrated biomechanical features in real time from sparse multiview images. In this section, we review the previous studies relevant to this research.

\subsection{Radiance field rendering}
Radiance field rendering-based techniques become popular in recent years because of the photorealistic scene representation capability.
NeRF~\cite{mildenhall2021nerf}, introduced in 2020, proposes a coordinate-based neural network to represent a 3D scene. 
The neural networks based on MLP map 3D coordinates and 2D view directions into density and color. 
Several follow-up works~\cite{barron2021mip,barron2022mip, park2021nerfies} are proposed and achieved impressive results, further verifying the capability.
In addition, further studies are conducted to address the limitations, including long training~\cite{ Dey2022B02, neff2021donerf, garbin2021fastnerf, lin2022efficient} and inference time\cite{muller2022instant, reiser2021kilonerf}, scene specificity~\cite{yu2021pixelnerf}, and static scene constraints~\cite{pumarola2021d, su2021nerf}.
In recent years, Gaussian splatting~\cite{kerbl20233d} techniques emerge as an alternative to implicit neural radiance fields by utilizing a set of 3D Gaussians to learn fast explicit scene representations.  

\subsection{Radiance field rendering for human}
Radiance field rendering techniques demonstrate promising results in various applications for 3D human representation. 
Early studies~\cite{zhao2022humannerf, xu2021h, su2021nerf} employ neural radiance fields to produce 3D human avatars from a spare set of images. 
Subsequent studies~\cite{hu2023sherf, chen2023gm, xu2021h, yu2023monohuman, jiang2023instantavatar} improve the generalizability of these models by incorporating the SMPL~\cite{SMPL:2015} parameters as input. 
Likewise, \cite{peng2023implicit, jiang2022neuman, weng2022humannerf, su2021nerf} utilize existing skeletal data, state-of-the-art pose estimators, or pose data to generate novel views and poses. 

Recently, thanks to the fast rendering speed of Gaussian splatting techniques, radiance field-based methods~\cite{zheng2023gps, hu2023gauhuman, kocabas2024hugs} have been able to represent 3D humans in real-time.
More recently, \cite{Dey_2024_CVPR} proposes generalized radiance fields for versatile human features that extend beyond RGB rendering by integrating additional human features, although the rendering speed is not real time.
Our approach stands out from prior works by combining generalizable Gaussian splatting techniques with feature splitting, which maintains the quality of both low- and high-frequency features, and a dedicated pose regression network to estimate 3D human avatars with integrated biomechanical features in real time.

\subsection{Human pose estimation}
Human features such as 3D pose and dense pose estimation are a long-standing problem in computer vision research. Many popular 2D and 3D pose estimation from images is based on supervised training. 
Algorithms for 2D pose estimation~\cite{OpenPose, fang2017rmpe, He_2017_ICCV, cheng2020bottom, Kreiss_2019_CVPR} employ 2D CNN architecture to estimate 2D poses from images. 
Works such as \cite{RogezWS18, ning2017knowledge, Moon_2019_ICCV_3DMPPE, Wang2018MagnifyNetFM} employ person detectors to estimate the poses of multiple persons. 
Bottom-up approaches like \cite{OpenPose, cheng2020bottom, Kreiss_2019_CVPR} identify joints using heatmaps and link body parts, but face challenges with occluded or partially visible body parts. 
Techniques for estimating 3D poses can be classified into direct methods and 2D-to-3D lifting methods. \cite{RogezWS18, Moon_2019_ICCV_3DMPPE} concentrate on determining 3D poses directly from images. 
In this paper, we introduce a new method for directly predicting 3D poses. 
\cite{guler2018densepose} introduce DensePose estimation from 2D images. 
Recent techniques \cite{guo2019adaptive, wang2020ktn} employ a multitask learning approach for DensePose estimation. 
In this study, we present a novel approach for learning DensePose estimation by employing the Gaussian splatting method. 
\section{Method}
We present HFGaussian, a unified framework that leverages Gaussian splatting to estimate human features and 3D pose in real-time. 

 \begin{figure*}[t]
\begin{center}
   \includegraphics[width=0.9\linewidth]{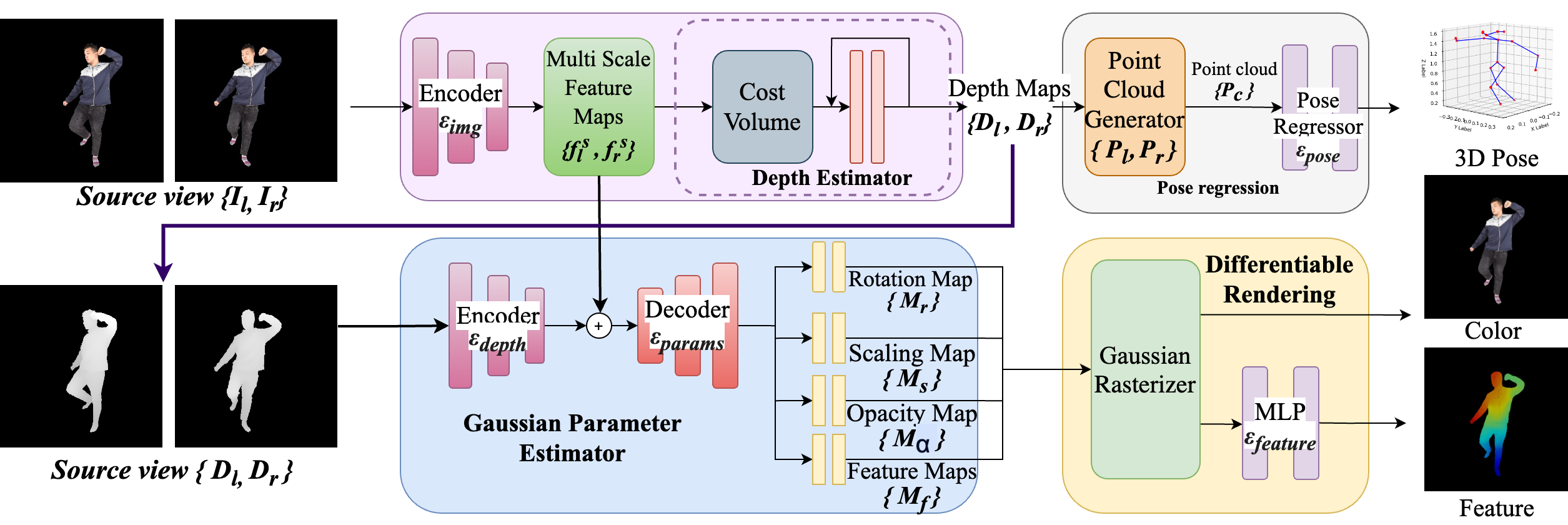}
\end{center}
\vspace{-10pt}
   \caption{
   \textbf{The HFGaussian pipeline:} Given a target view, the nearest source views $I_l$ and $I_r$ are selected, and passed through an image encoder $\epsilon_{img}$ to generate feature maps $f_l^s$ and $f_r^s$ for depth maps $D_l$ and $D_r$ estimation. 
   The depth maps are then encoded using a $\epsilon_{depth}$ encoder and combined with the image features before passing through a U-Net based decoder $\epsilon_{params}$ to predict Gaussian feature maps $\mathcal{M}_r$, $\mathcal{M}_s$, $\mathcal{M}_\alpha$, and $\mathcal{M}_f$. 
   Finally, the predicted Gaussians are splatted and rasterized to generate the novel view and human features, which are further processed by a smaller MLP $\epsilon_{feature}$ to obtain the final human features.}
\label{fig:method}
\end{figure*}

\subsection{Preliminary}
\label{prelim}
In this study, we present a novel method for reconstructing the 3D human model while incorporating biomechanical characteristics through the use of Gaussian splatting. 
Here, we briefly outline the fundamental concepts behind 3D Gaussian splatting~\cite{kerbl20233d} and generalizable Gaussian splatting~\cite{zheng2023gps} techniques. 
3DGS is introduced as an alternative explicit representation in contrast to continuous NeRF-based approaches. 
A scene in 3DGS is represented through a collection of 3D Gaussians characterized by certain properties: the 3D position of Gaussian $\mu$, color $c$ represented using spherical harmonics, opacity $\alpha$, and a 3D covariance matrix $\Sigma$. 
A 3D Gaussian in 3DGS is represented as:
\[ G(x) = e^{-\frac{1}{2}(x-\mu)^T\Sigma^{-1}(x-\mu)}\]
The covariance matrix $\Sigma$ can be broken down into a rotation matrix $\textbf{R}$ and a scaling matrix $\textbf{S}$. 
The 3D Gaussians are projected into 2D space using a view transformation matrix $\textbf{W}$ and Jacobian of an affine approximation of the projective transform $\textbf{J}$. The 2D covariance matrix $\Sigma’$ is represented as $\Sigma’ = \textbf{JW}\Sigma\textbf{W}^T\textbf{J}^T$. The alpha blending technique similar to NeRF, is used to rendering final pixel colors from Gaussians:
\[C = \sum_{i \in \mathcal{N}} c_i \alpha'_i \prod_{j=1}^{i-1} (1 - \alpha'_j)\]
where $c_i$ represents the learned color, $\alpha_i’$ denotes the result of the multiplication between the opacity $\alpha_i$ and the 2D Gaussian.
The 3DGS technique demonstrates notably faster rendering speeds compared to continuous methods based on NeRF, primarily because it can directly project and blend 3D Gaussian into color.

Although 3DGS is efficient and produces high-quality results, vanilla 3DGS is scene-specific and is not generalizable to new scenes. 
To address this issue, GPS-Gaussian~\cite{zheng2023gps} proposes Gaussian parameter maps on the source views and directly estimates instant novel views. 
They focus only on human subjects and train on a large amount of human data. 
Given sparse source views and a novel target view $I_{tar}$, GPS-Gaussian selects 2 neighboring views $I_r$ and $I_l$ of the target view. 
The source views are then passed through an image encoder $\varepsilon_{img}$ to extract dense feature maps $f^s \in \mathbb{R}^{H/2^s \times W/2^s \times D_s }$, corresponding to each source image. 
Using the feature maps from each source view ($f_r^s$, $f_l^s$), a 3D correlation volume $C$ is generated. 
This correlation volume along with the corresponding camera parameters for the source views ($K_r$, $K_l$) is used to iteratively estimate depth maps. 
It can be formulated as: \[<\mathbf{D}_l, \mathbf{D}_r> = \phi_{depth}(f_l^s, f_r^s, K_l, K_r),\] where $\phi_{depth}$ represents the depth estimation module. 
Later, these depth estimations are used to generate the position of 3D Gaussians. 

To predict the Gaussian parameters, they employ a mapping function that formulates the 3D Gaussian from the 2D image plane. 
When given a foreground pixel coordinate $x$ in the image plane, the Gaussian map can be represented as:
\[\mathbf{G}(x) = \{\mathcal{M}_p(x), \mathcal{M}_c(x), \mathcal{M}_r(x), \mathcal{M}_s(x), \mathcal{M}_\alpha(x)\}\]
The previously estimated depth and camera projection matrix can be utilized to unproject the pixel coordinates $x$ from the image plane to the 3D coordinates, represented as $\mathcal{M}_p(x)$. 
The color map uses the source image color directly: $\mathcal{M}_c(x) = I(x)$. 
To estimate the remaining Gaussian parameters, they first generate depth features using a depth encoder. 
These features are then combined with the image feature and passed through a U-Net like decoder to generate pixel-wise Gaussian features: \[\Gamma = D_{parm}(\epsilon_{img}(I) \oplus \epsilon_{depth}(I))\], where $\oplus$ represents the concatenation operation.
Finally, three different prediction heads are used to generate the remaining Gaussian feature maps $\mathcal{M}_r$, $\mathcal{M}_s$, and $\mathcal{M}_\alpha$ represent the rotation, scaling, and opacity head, respectively. 
The entire method is end-to-end differentiable and optimized using photometric and depth loss. 

\subsection{Learning human feature with 3DGS}
\label{feature}
We propose a novel method for real-time 3D human avatar estimation with biomechanic properties. 
Although recent advancements in 3D Gaussian splatting techniques~\cite{kerbl20233d} show remarkable efficiency in 3D scene representation compared to previous approaches, the vanilla method still requires per-subject optimization for scene representation.
To address this limitation, GPS-Gaussian~\cite{zheng2023gps} introduces a generalizable approach for 3D human representation that takes advantage of 2D Gaussian parameter maps to estimate 3D Gaussians. 
This approach enables the direct regression of 3D Gaussian properties, facilitating instant novel view synthesis without the need for fine-tuning or optimization. 

Building on this, we propose to extend the capabilities of 3D Gaussian splatting by learning a generalizable representation of 3D humans with integrated biomechanic properties. 
We develop a novel architecture that can predict 3D pose and human features along with photometric and geometric representation for novel views. 
Similarly to GPS-Gaussian, given a target view, we select the two nearest source views, $I_l$ and $I_r$, which are RGB images ($H\times W$) corresponding to the left and right views. 
These two views are then fed into a shared image encoder $\epsilon_{img}$ to generate multiscale image features $f_l^s$ and $f_r^s$ where $s$ is the feature scale.

From the feature maps $(f_l^s, f_r^s)$ of each source view, a cost volume $C$ is generated by correlating the two feature maps. 
Then, an iterative update mechanism is used to estimate depth maps $(D_l, D_r)$ corresponding to each source view. 
These estimated depth maps are subsequently used as input to the pose regression network and Gaussian parameter estimator, which are discussed in Sec. \ref{pose} and \ref{dense}.
The pose regression network is capable of outputting the 3D pose of human subjects.

The scene is represented using a set of optimized 3D Gaussians, where each Gaussian is characterized by $G=\{X, c, r, s, \alpha, f\}$ where $X$ denotes the 3D position, $c$ represents the color, $r$ signifies the rotation, $s$ corresponds to the scaling, $\alpha$ indicates the opacity, and $f$ encodes the human feature. 
We estimate the 3D Gaussian parameters on 2D planes using Gaussian parameter maps represented as:
\[G(x) = \{\mathcal{M}_p(x), \mathcal{M}_c(x), \mathcal{M}_r(x), \mathcal{M}_s(x), \mathcal{M}_\alpha(x), \mathcal{M}_f(x)\} \]
where $x$ denotes the coordinates of a foreground pixel within the image plane, $\mathcal{M}_p$, $\mathcal{M}_c$, $\mathcal{M}_r$, $\mathcal{M}_s$, $\mathcal{M}_\alpha$, and $\mathcal{M}_f$ represent the Gaussian parameter maps corresponding to position, color, rotation, scaling, opacity, and feature, respectively. 
The $\mathcal{M}_p$ function maps the 2D image pixel coordinates $x$ to the 3D space by using the predicted depth information and the known camera parameters. 
The $\mathcal{M}_c$ function directly uses the RGB color values from the source images.

To estimate the remaining four Gaussian parameters, we use an encoder $\epsilon_{depth}$ to encode depth maps and then employ a U-Net-like decoder $\epsilon_{params}$ to generate a Gaussian feature $\Gamma$ from depth encoding and image encoding. 
Finally, we use four separate prediction heads implemented with convolutional layers to estimate Gaussian parameters. 
The prediction head for the human feature is defined as  $\mathcal{M}_f = Sigmoid(h_f(\Gamma(x)))$ , where $h_f$ represents the feature head.

\subsection{Pose regression Network}
\label{pose}
 \begin{figure}[h!]
\begin{center}
   \includegraphics[width=1\linewidth]{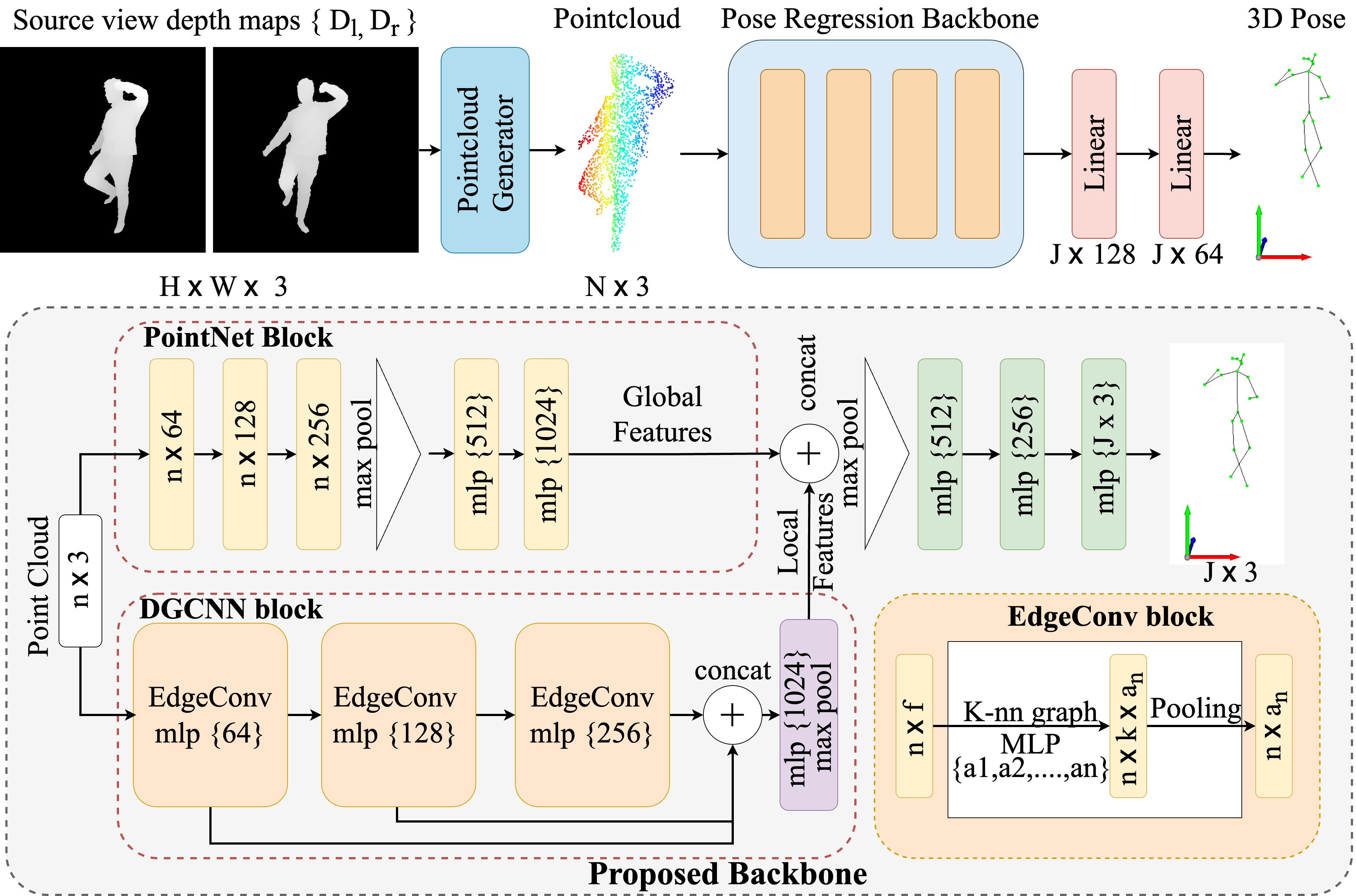}
\end{center}
\vspace{-10pt}
   \caption{
   \textbf{Pose Regression Network Overview:} The network takes point clouds generated from depth maps as input and outputs 3D poses. We compare three point cloud classification backbones and propose a novel architecture combining PointNet and DGCNN architecture for robust feature extraction.}
   
\label{fig:poseregress}
\end{figure}
A key component of the architecture is the pose regression network, which aims to estimate the 3D pose of the human subject from the partial point cloud data generated from depth maps $(D_l, D_r)$, also referred to as the subset of the 3D Gaussians. 
Estimation of 3D human keypoints from point cloud data is an active research area, with prior work exploring various approaches to address these challenges~\cite{zanfir2023hum3dil, zhou2020learning, chen2022efficient}. 
As mentioned in the previous section, the depth estimation models predict depth maps $(D_l, D_r)$ corresponding to each source view $(I_l, I_r)$. 
These depth maps of the source views and their associated camera parameters can be used to generate point clouds $P_l, P_r$. 
The 2D masks of each source view are then used to extract only the point cloud of the human. 
The point clouds from both views are then combined as they are in the same 3D frame to generate a combined point cloud. 
From this combined point cloud, 2048 points are randomly sampled, and then this sampled point cloud $P_c$ is used as input to the pose regression network. 
For the pose regression network, we experiment with three different popular backbone architectures, namely PointNet~\cite{qi2017pointnet}, DGCNN~\cite{wang2019dynamic}, and Point Transformer~\cite{zhao2021point}, which are well known for point cloud classification. 
In this work, we compare the performance of all three models on 3D and 2D pose estimation, using their corresponding classification backbone with added MLP layers at the end for pose estimation, as shown in figure~\ref{fig:poseregress}. 
The PointNet architecture used in this work is a customized version based on the implementation described in \cite{chen2022efficient}. 
In this work, we introduce a novel pose regression backbone that combines global features from PointNet architecture and local features from DGCNN architecture while maintaining computational efficiency similar to PointNet for real-time inference. 
The proposed architecture for the pose regression network is shown in Figure~\ref{fig:poseregress}. 
The proposed architecture achieves comparable performance to the more complex Point Transformer network yet remains as efficient as PointNet, making it suitable for our real-time applications. 
As the proposed model learns both global and local features jointly, it can provide a more robust pose estimation that is resilient to noisy and incomplete point cloud data.

\subsection{Human feature estimation}
\label{dense}
The proposed method also estimates human features. 
We demonstrate the capabilities of our approach by estimating dense pose, which involves predicting Continuous Surface Embeddings for the human subject. 
We include an additional branch in the Gaussian parameter estimator to predict human feature maps $\mathcal{M}_f$ for each Gaussian, similar to the rotation and opacity maps. 
Inspired by feature splatting~\cite{martins2024feature}, we apply a similar technique that estimates human feature vectors $f_p$ by splatting Gaussian features $f_i$ in the image plane, and then blending the feature vectors using alpha composition: 
\[f_p = \sum_{i \in \mathcal{N}} f_i \alpha'_i \prod_{j=1}^{i-1} (1 - \alpha'_j)\]
The blended feature vectors $f_p$ are decoded by a MLP consisting of two linear layers with ReLU activation functions, followed by a final layer with a sigmoid activation function, to render the continuous surface embeddings.

\subsection{Optimization}
The proposed HFGaussian method comprises three key components: generalizable Gaussian splatting, 3D pose estimator, and human feature estimator module. 
The model is trained in two stages. 
First, the depth estimator module is trained on both source views. 
Then, the Gaussian parameter estimator module is trained using the depth maps and image features, along with the feature estimator MLP and the 3D pose estimator. 
A combined loss function is utilized to train all three parts simultaneously, as described:
\[ \mathcal{L} = \mathcal{L}_{image} +  \mathcal{L}_{depth} + \mathcal{L}_{pose} + \mathcal{L}_{feature} \]
where $\mathcal{L}_{image}$ is the photometric loss between the ground truth and the rendered image represented as $\mathcal{L}_{image} = \beta \mathcal{L}_{mae} + \gamma \mathcal{L}_{ssim}$ where $\beta$ and $\gamma$ are 1.6 and 0.4 respectively. 
The depth loss $\mathcal{L}_{depth}$ is defined as: \[
\mathcal{L}_{\text{depth}} = \sum_{t=1}^{T} \mu^{T-t} \| \mathbf{d}_{gt} - \mathbf{d}^{t} \|_1
\] where $d$ represents the depth and $\mu$ is set to 0.9 for our experiments. 
The 3D pose estimation loss $\mathcal{L}_{pose}$ is the L2 loss between ground truth and estimated 3D keypoints. 
Lastly, the feature estimation loss $\mathcal{L}_{feature}$ is the L1 loss between the ground truth and the predicted human features, which in this case are the continuous surface embeddings.

\section{Experimental Results}
We conduct a comprehensive evaluation of our model's ability to learn a generalized Gaussian representation of humans, including both 3D human poses and features.
Extensive experiments are conducted on a variety of datasets, and we evaluate our results with other state-of-the-art NeRF and Gaussian splatting-based methods.
\subsection{Implementation details}
The proposed HFGaussian method is implemented using the PyTorch framework and the AdamW optimizer~\cite{loshchilov2017decoupled} with a learning rate of $2 \times 10^{-4}$ and a weight decay of $1 \times 10^{-5}$. 
First, the depth estimation module is trained for 40,000 iterations, and then all three components of the network are jointly trained for 100,000 iterations with a batch size of 4 for all experiments. 
The complete training process takes approximately 14 hours on the dataset proposed in this study. 
For all experiments, we utilize the official versions of GHNeRF~\cite{Dey_2024_CVPR} and ENeRF~\cite{lin2022efficient}, training them for 100,000 iterations. Details about the evaluation metrics can be found in the supplementary Section \ref{metrics}.

\subsection{Dataset}
In this work, we create a custom dataset similar to the one utilized in GPS-Gaussian \cite{zheng2023gps}. 
The GPS-Gaussian dataset is created using 526 scans from the THuman2.0 \cite{tao2021function4d} dataset and includes images, masks, depth information, and camera parameters. 
The dataset is insufficient for this study and required additional ground-truth data on various human features, such as keypoints and dense pose, in order to train the proposed method. 
To this extent, we extend the GPS-Gaussian dataset by generating additional ground-truth information. 
We start with 526 human scans and SMPLX~\cite{SMPL:2015} parameters from THuman2.0. 
Next, we employ the SMPLX model fitted to the scan to produce accurate 3D poses. 
We produced 19 key points representing the main joints of the human body (\textit{nose, neck, right shoulder, right elbow, right wrist, left shoulder, left elbow, left wrist, pelvis, right hip, right knee, right ankle, left hip, left knee, left ankle, right eye, left eye, right ear, left ear}). 
To produce the 2D keypoints, we map the 3D keypoints into image coordinates utilizing the camera's parameters. 
We use DensePose \cite{guler2018densepose} to generate ground-truth Continuous Surface Embeddings for each image in the dataset. 
To render images from human scans, we follow the same convention of GPS-Gaussian, where 8 images are rendered in a circle around the human at 45-degree intervals, and those are used as source views. 
Additionally, 3 random viewpoints are generated as target views. 
We generate all images in the dataset at a resolution of $512 \times 512$.
We also use the real-world dataset captured by \cite{zheng2023gps} to evaluate our methods. 
This real dataset does not provide any ground-truth 3D or 2D keypoint information. 
To further evaluate the generalization ability of our model, we also preprocess and utilize the THuman4.0 dataset~\cite{zheng2022structured}, which contains three clips of real-world subjects, resulting in test sets with 19656, 40464, and 24880 samples, respectively.
We will release the preprocessed THuman2.0 and THuman4.0 after the paper is published.

\subsection{Baseline} 
The proposed HFGaussian method is compared with other state-of-the-art generalizable approaches for human subjects. 
Regarding the benchmarking of human features and pose estimation, many previous generalizable techniques lack the ability to estimate human features and pose simultaneously. 
We select GHNeRF~\cite{Dey_2024_CVPR} as a baseline, as it can estimate human features such as 2D keypoints and dense pose, although not in real time.

\subsection{Novel view synthesis}
Our proposed methods are compared with other state-of-the-art generalizable radiance field rendering techniques for human representation. 
\begin{figure}[h!]
    \centering
    \includegraphics[width=1\linewidth]{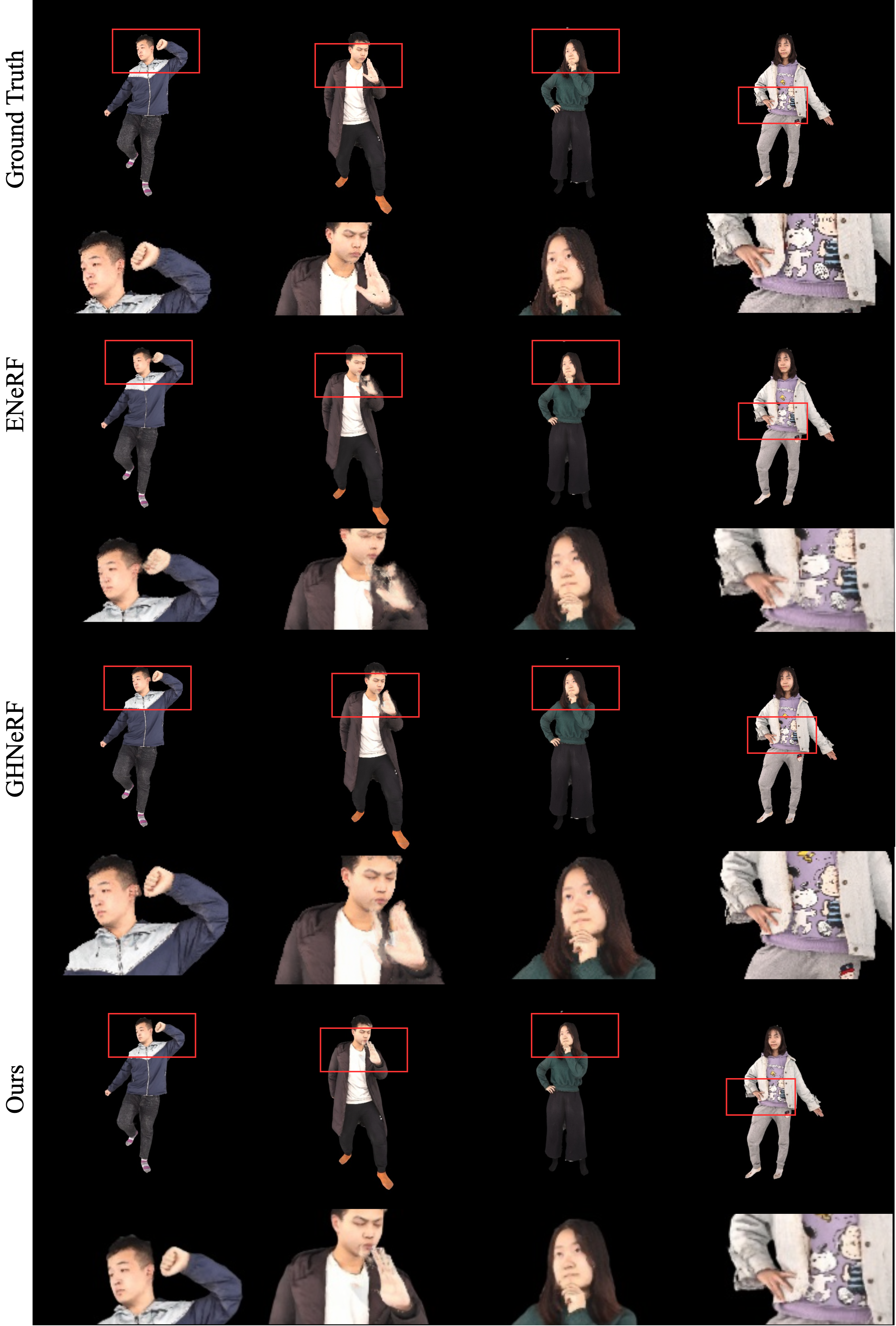}
    \caption{Qualitative comparison of novel view synthesis results on THuman2.0 test set.}
    \label{fig:novel_res}
\end{figure}
The quantitative results for novel view synthesis presented in Table~\ref{tab:compare_method} indicate that our methods are competitive with other state-of-the-art approaches while also estimating human 3D joints and poses, going beyond mere RGB image synthesis.
Among all the methods evaluated, GPS-Gaussian achieves the highest PSNR of 32.55. 
Our method closely matches this performance, achieving a PSNR of 32.43, while also estimating other human features simultaneously. 
The qualitative results for novel view synthesis are shown in Figure~\ref{fig:novel_res}. 
\begin{table*}[h!]
    \centering
    \scalebox{1}{
    \small
    \begin{tabular}{>{\raggedright\arraybackslash}p{3cm}|ccc|c|c|c|c}
    \hline
    & & Image & & Dense Pose & 3D Pose & 2D Pose & Inference Time\\
    \hline
         Method & PSNR $\uparrow$ & SSIM $\uparrow$ & LPIPS $\downarrow$ & MSE $\downarrow$ & MPJPE $\downarrow$ & PCK $\uparrow$  & FPS $\uparrow$\\
    \hline
        ENeRF~\cite{lin2022efficient} & 32.51 & 0.9823 & 0.0245 & - & - & - & 28.97 \\
        GPS-Gaussian~\cite{zheng2023gps} & 32.55 & 0.9737 & 0.0300 & - & - & - & 32.02 \\
        \hline
        GHNeRF~\cite{Dey_2024_CVPR} & 32.98 & 0.9839 & 0.0210 & 0.0020 & - & 0.6767 & 11.22 \\
       \rowcolor{yellow} Ours & 32.43 & 0.9734 & 0.0303 & 0.0017 & 0.0704 & 0.8707 & 24.37 \\
    \hline
    \end{tabular}}
    \caption{Quantitative results for novel view synthesis on the THuman2.0 dataset. Photometric quality is assessed via PSNR, SSIM, and LPIPS metrics, while dense pose estimation accuracy is measured using mean square error (MSE).}
    \label{tab:compare_method}
\end{table*}

\begin{figure}[h!]
    \centering
    \includegraphics[width=1\linewidth]{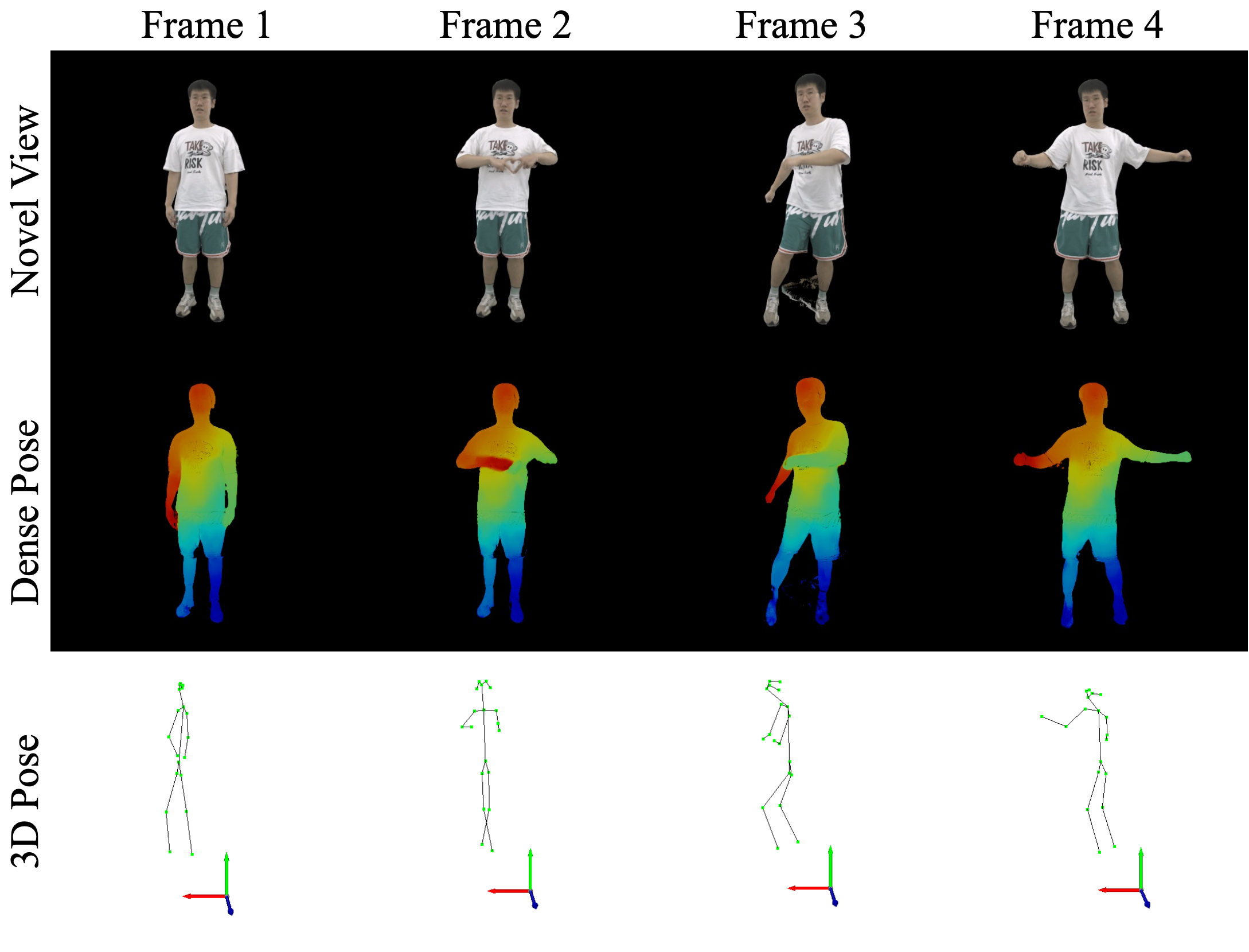}
    \caption{Qualitative result of the proposed method of real human dataset. The figure illustrates random frames from real human data and their corresponding novel view, dense pose, and 3D pose predicted by the HFGaussian. The 3D pose images are generated by projecting the 3D pose in a 2D plane from a fixed viewing angle.}
    \label{fig:real_data}
\end{figure}
To demonstrate the generalization capability of the proposed method, we evaluate our pre-trained model directly on the real-world dataset by \cite{zheng2023gps} and THuman4.0 real-world dataset without any fine-tuning.
The qualitative results present in Figure~\ref{fig:real_data} and Figure~\ref{fig:th4_res} demonstrate the robustness of our approach, generalizing unseen data in real time and accurately estimating biomechanical features.
The quantitative results in Table~\ref{tab:compare_method_th4} demonstrate that our approach is much more robust across datasets and outperforms all baselines.
Furthermore, the results indicate that our method can reliably estimate biomechanical features even in challenging and self-occlusion scenarios in real time.

\begin{table*}[h!]
    \centering
    \scalebox{1}{
    \small
    \begin{tabular}{>{\raggedright\arraybackslash}p{6cm}|ccc|c|c|c}
    \hline
        & & Image & & Dense Pose & 3D Pose & 2D Pose\\
    \hline
         Method & PSNR $\uparrow$ & SSIM $\uparrow$ & LPIPS $\downarrow$ & MSE $\downarrow$ & MPJPE $\downarrow$ & PCK $\uparrow$ \\
    \hline
        Ours + PointNet + 3D pose & 32.4087 & 0.9730 & 0.0303 & 0.00176 & 0.0927 & -\\
        Ours + DGCNN + 3D pose & 32.4131 & 0.9732 & 0.0303 & 0.00174 & 0.0811 & -\\
        Ours + PointTransfomer + 3D pose & 32.4141 & 0.9730 & 0.0303 & 0.00174 & 0.0894 &  -\\
        \rowcolor{yellow}Ours + Proposed pose regressor + 3D pose & 32.4300 & 0.9734 & 0.0303 & 0.00173 & 0.0704 & -\\
        \hline
        Ours + PointNet + 2D pose & 32.4300 & 0.9728 & 0.0311 & 0.00175 & - & 0.8309 \\
        Ours + DGCNN + 2D pose & 32.4020 & 0.9731 & 0.0304 & 0.00173 & - & 0.8707\\
        Ours + PointTrans + 2D pose & 32.4000 & 0.9729 & 0.0304 & 0.00177 & - & 0.7721 \\
        \rowcolor{yellow}Ours + Proposed pose regressor + 2D pose & 32.4300 & 0.9731 & 0.0304 & 0.00176 & - & 0.8680 \\
    \hline 
    \end{tabular}}
    \caption{The table shows quantitative results of our approach on the Thuman2.0 dataset. The results are categorized into four groups: novel view, dense pose estimation, 3D pose estimation, and 2D pose estimation.}
    \label{tab:hfg}
\end{table*}

\subsection{3D/2D Pose Estimation}
The proposed HFGaussian method estimates 3D keypoints using a pose regression network, which takes the point cloud generated by the position of 3D Gaussians as input. 
In contrast, many state-of-the-art approaches do not provide 3D keypoints or any human biomechanical features. 
Our method is the first to directly estimate 3D pose without relying on prior supervision or parametric body models. 
The quantitative results of our approach are presented in Table~\ref{tab:hfg}, and the qualitative results are shown in Figure~\ref{fig:3d_pose}. 
As mentioned earlier, we experiment with different backbone architectures for the pose regression network. 
We then propose a novel architecture that combines elements of PointNet~\cite{qi2017pointnet} and DGCNN~\cite{wang2019dynamic}. 
The results demonstrate that the proposed backbone architecture achieves the best MPJPE score among all backbones tested, as it effectively combines global and local features to provide robust 3D pose estimation. 
This is also evident in the visual comparison presented in Figure~\ref{fig:3d_pose}.

In addition to 3D pose estimation, we also evaluate 2D pose estimation using the same pose regression network, with a modified final linear layer for 2D keypoint prediction. 
We compare our 2D pose estimation performance with that of GHNeRF~\cite{Dey_2024_CVPR}, which is also capable of estimating 2D keypoints. 
The quantitative results are presented in Table~\ref{tab:compare_method}, and the qualitative comparisons are shown in Figure~\ref{fig:keypoint_res}. 
The experiments demonstrate that our method significantly outperforms GHNeRF in the estimation of 2D key points.

\begin{figure}[h!]
    \centering
    \includegraphics[width=1\linewidth]{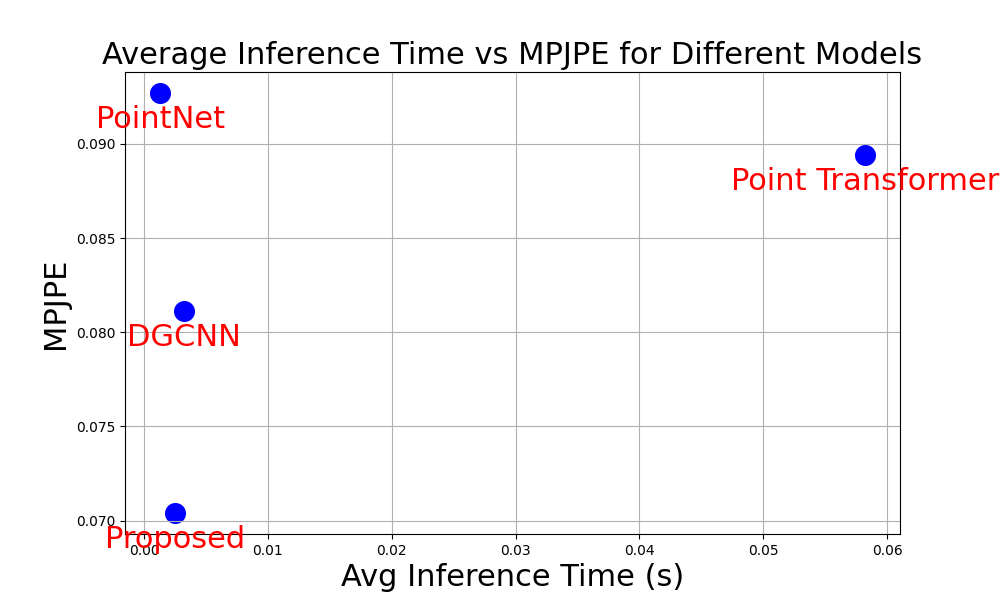}
    \caption{The figure illustrates the performance and average inference time of different 3D pose estimation backbone architectures. The 3D pose estimation performance is measured in terms of MPJPE, which is plotted on the y-axis. The x-axis represents the average inference time of the respective backbone models. 
    }
    \label{fig:model_performance}
\end{figure}
\subsection{Dense Pose Estimation}
The proposed HFGaussian method is capable of estimating various human features using the feature splatting technique discussed previously. 
To showcase the feature prediction capabilities of the proposed approach, we conduct a dense pose estimation task as an illustrative example. 
Specifically, we estimate the Continuous Surface Embedding to represent the dense pose. 
The quantitative results, presented in Table~\ref{tab:compare_method}, compare the dense pose estimation performance of our method against the GHNeRF baseline. 
Furthermore, the qualitative results, depicted in Figure~\ref{fig:dense_res}, further demonstrate that our method outperforms GHNeRF in learning the dense pose representation. 
Additionally, we showcase our model's ability to estimate dense pose on real-world data, as illustrated in Figure~\ref{fig:real_data}.

\subsection{Real Time Performance}
One of the primary objectives of this work is to achieve real-time performance during inference and estimate novel views with biomechanical features on unseen data. 
We have compared the real-time performance of our method with GPS-Gaussian~\cite{zheng2023gps} and GHNeRF~\cite{Dey_2024_CVPR}, among which only GHNeRF is capable of estimating additional human features such as 2D keypoints. 
The results presented in Table~\ref{tab:rendering_speed} demonstrate the performance of different methods in terms of frames per second. 
In our method, we experiment with various backbones for 3D pose estimation. 
The relationship between the different backbones, the inference speed, and the precision is shown in Figure~\ref{fig:model_performance}. 
The figure indicates that the proposed architecture for the 3D pose estimation backbone achieves the best balance between speed and accuracy.
\begin{table}[h]
    \centering
    \small
    \begin{tabular}{l|r}
    \hline
         Method & FPS \\
    \hline
         GHNeRF + ResNet & 11.22\\
         GHNeRF + DINO & 4.08\\
         GPS-Gaussian & 32.02 \\
         Ours + PointNet & 25.07\\
         Ours + DGCNN & 24.18\\
         Ours + Point Transformer & 10.29 \\
         \rowcolor{yellow}Ours + Proposed & 24.37\\
    \hline
    \end{tabular}
    \caption{Average rendering speed in FPS(Frame per second). GPS-Gaussian represent the baseline method.}
    \label{tab:rendering_speed}
\end{table}
\subsection{Ablation Studies}
\begin{figure}[h]
    \centering
    \includegraphics[width=1\linewidth]{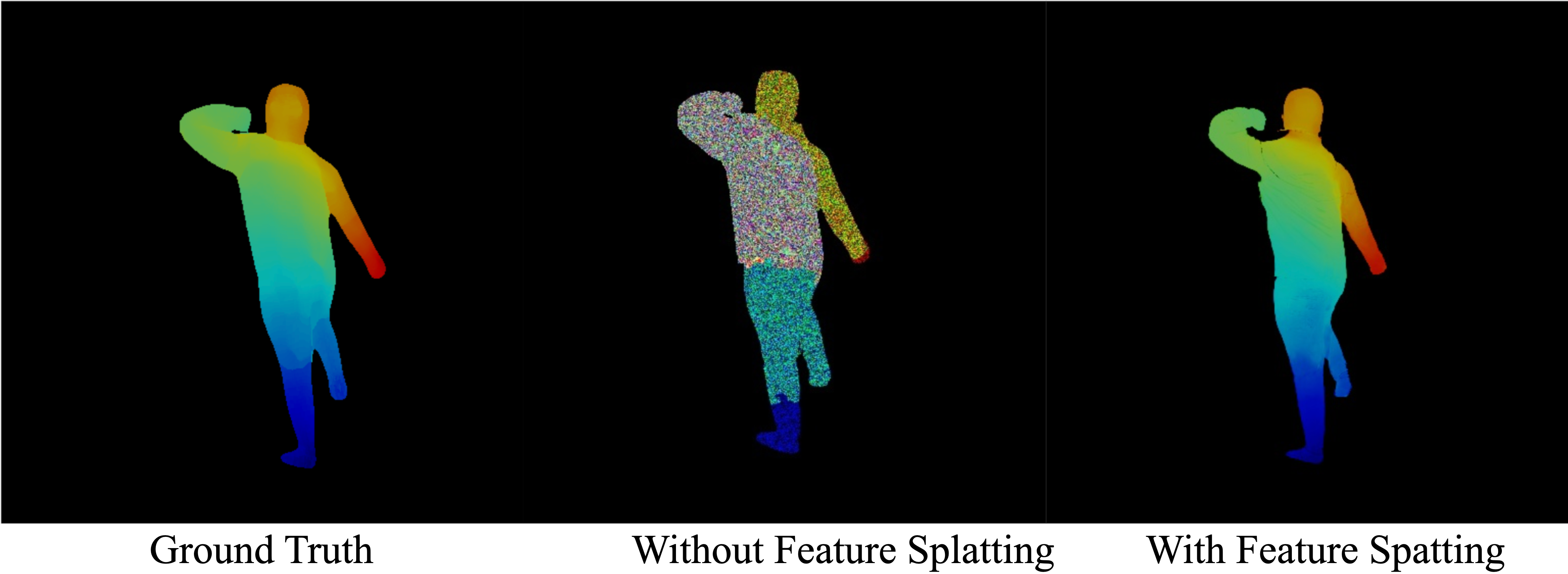}
    \caption{The significance of feature splatting methods in estimating human features.}
    \label{fig:featsplat}
\end{figure}

To assess the importance of feature splatting for estimating human features, we conduct an ablation study. 
We compare the performance of our method in estimating dense pose, a key human feature, using feature splatting versus directly predicting feature values similar to RGB color. 
In both cases, we keep the other Gaussian parameters constant. 
The qualitative results presented in Figure~\ref{fig:featsplat} demonstrate that the method without feature splatting is unable to accurately estimate dense pose, as the same Gaussians are unable to effectively represent both RGB values and feature attributes simultaneously.
\vspace{-10pt}
\section{Conclusion}
This paper presents a novel framework, \shortnamenospace, for the real-time rendering of human avatars with biomechanical properties. 
The proposed method utilizes Gaussian splatting to generate novel views from sparse source views of human subjects in real time. 
Extensive experimentation demonstrates the effectiveness of this approach, which represents a significant improvement over previous Gaussian splatting-based methods for human representation. 
While the primary focus of this work is on 3D human pose estimation and feature splatting-based dense pose estimation, we believe that feature splatting can be leveraged to learn other human-centric features, such as body part segmentation. 
In summary, this work presents promising results and opens up new avenues for research in 3D human representation.



{\small
\bibliographystyle{ieee_fullname}
\bibliography{main}
}

\pagebreak
\clearpage
\normalsize
\twocolumn[
\begin{center}
\textbf{\large Supplementary Material: \longname}
\end{center}
]

\setcounter{equation}{0}
\setcounter{figure}{0}
\setcounter{table}{0}
\setcounter{page}{1}
\setcounter{section}{0}
\makeatletter
\renewcommand{\theequation}{S\arabic{equation}}
\renewcommand{\thefigure}{S\arabic{figure}}
\renewcommand{\thetable}{S\arabic{table}}
\newcommand{\xmark}{\ding{55}}
\newcommand{\NA}{---}

\section{Author Statement}
In this supplementary material, we present information regarding the preprocessed dataset, evaluation metrics, additional visualization results, and analysis. We are committed to the ongoing maintenance and support of the preprocessed dataset. The dataset is distributed under an MIT license, allowing use, redistribution, and citation in accordance with the license terms.

\section{Dataset}

In this work, we extend the GPS-Gaussian dataset, which includes 526 scans from the THuman2.0 dataset, by generating additional ground-truth data, such as 3D poses, 2D keypoints, and DensePose embeddings. 
Starting with SMPLX parameters fitted to the scans, we extract 19 keypoints representing major human joints and project the 3D keypoints into 2D image coordinates using the camera parameters. 
Images are rendered from 8 source views at 45-degree intervals around each scan, along with 3 random target viewpoints, all at a resolution of 512x512. 
Additionally, we preprocess the THuman4.0 dataset, which includes three clips of real-world subjects captured by 24 cameras arranged in a circle. 
This results in test sets containing 19,656, 40,464, and 24,880 samples, where the two source views of each sample are also separated by 45 degrees.
The key difference between our THuman2.0 and THuman4.0 datasets is that THuman4.0 is based on real-world RGB images with real-world camera poses, which may not form a perfect circle.
This difference can verify the generalization ability of the proposed method (see supplementary Section~\ref{add_res}).

Table~\ref{table:supp_dataset} demonstrates that our THuman2.0 and THuman4.0 datasets cover a wide range of human features that are lacking in existing datasets. 
We plan to release our datasets after publication to encourage further development in this field.

\begin{table*}[tb]
\centering
\begin{tabular}
{l|cccccc}
\hline
Dataset & Multi-View  & RGB & 2D Keypoints & 3D Keypoints & Dense Pose & Real \\ 
\hline
LSP~\cite{johnson2010clustered} & \xmark & \checkmark & \checkmark & \xmark & \xmark & \checkmark \\
MPII~\cite{andriluka14cvpr} & \xmark & \checkmark & \checkmark & \xmark & \xmark & \checkmark \\
Smplify-X~\cite{SMPL-X:2019}  & \xmark & \checkmark & \checkmark & \checkmark & \checkmark & \checkmark \\
THuman2.0~\cite{tao2021function4d} & \checkmark & \checkmark & \xmark & \xmark & \xmark & \xmark \\ 
THuman4.0~\cite{zheng2022structured} & \checkmark & \checkmark & \xmark & \xmark & \xmark & \checkmark \\ 
Human3.6M~\cite{h36m_pami} & \checkmark & \checkmark & \checkmark & \checkmark & \xmark & \xmark \\ 
SURREAL~\cite{varol17_surreal} & \checkmark & \checkmark & \checkmark & \checkmark & \checkmark & \xmark \\
3DPW~\cite{vonMarcard2018} & \checkmark & \checkmark & \checkmark & \checkmark & \xmark & \checkmark \\
ZJU-MoCap~\cite{peng2021neural} & \checkmark & \checkmark & \checkmark & \checkmark & \xmark & \checkmark \\ 
\rowcolor{yellow}THuman2.0 (Ours) & \checkmark & \checkmark & \checkmark & \checkmark & \checkmark & \xmark \\ 
\rowcolor{yellow}THuman4.0 (Ours) & \checkmark & \checkmark & \checkmark & \checkmark & \checkmark & \checkmark \\ 
\hline
\end{tabular}
\caption{Comparison between our preprocessed dataset and other human-centric datasets. In this table, Real denotes high-quality, realistic images of actual human scenes captured by physical cameras instead of virtual cameras.}
\label{table:supp_dataset}
\end{table*}

\begin{table*}[tb]
    \centering
    \scalebox{1}{
    \small
    \begin{tabular}{>{\raggedright\arraybackslash}ll|ccc|c|c|c}
    \hline
        & & & Image & & Dense Pose & 3D Pose & 2D Pose\\
    \hline
         Subject & Method & PSNR $\uparrow$ & SSIM $\uparrow$ & LPIPS $\downarrow$ & MSE $\downarrow$ & MPJPE $\downarrow$ & PCK $\uparrow$ \\
    \hline
        \multirow{4}{*}{S00} & ENeRF~\cite{lin2022efficient} & 31.121 & 0.978 & 0.029 & - & - & - \\
        & GPS-Gaussian~\cite{zheng2023gps} & 32.505 & 0.977 & 0.0196 & - & - & - \\
        & GHNeRF~\cite{Dey_2024_CVPR} & 29.117 & 0.969 & 0.049 & - & - & 0.3714 \\
        \rowcolor{yellow}& Ours & 33.610 & 0.981 & 0.017 & 0.0051 & 0.104 & 0.0018 \\
    \hline
        \multirow{4}{*}{S01} & ENeRF~\cite{lin2022efficient} & 29.204 & 0.966 & 0.041 & - & - & - \\
        & GPS-Gaussian~\cite{zheng2023gps} & 30.324 & 0.966 & 0.0272 & - & - & - \\
        & GHNeRF~\cite{Dey_2024_CVPR} & 27.805 & 0.962 & 0.050 & - & - & 0.4241 \\
        \rowcolor{yellow}& Ours & 30.968 & 0.968 & 0.026 & 0.0065 & 0.119 & 0.0029 \\
    \hline
        \multirow{4}{*}{S02} & ENeRF~\cite{lin2022efficient} & 27.993 & 0.968 & 0.044 & - & - & - \\
        & GPS-Gaussian~\cite{zheng2023gps} & 28.560 & 0.967 & 0.0324 & - & - & - \\
        & GHNeRF~\cite{Dey_2024_CVPR} & 26.165 & 0.964 & 0.050 & - & - & 0.3981 \\
        \rowcolor{yellow}& Ours & 29.051 & 0.967 & 0.033 & 0.0109 & 0.143 & 0.0041 \\
    \hline
    \end{tabular}}
    \caption{Quantitative results for novel view synthesis on the THuman4.0 dataset.}
    \label{tab:compare_method_th4}
\end{table*}

\section{Metrics} \label{metrics}
We use six distinct metrics to evaluate the quality of the predicted RGB images, 3D pose, and dense pose estimation.
For RGB image reconstruction, we use the peak signal-to-noise ratio (PSNR) and Structural Similarity Index (SSIM) to compare the quality, with higher values indicating better quality. 
Additionally, the Learned Perceptual Image Patch Similarity (LPIPS)~\cite{zhang2018perceptual} metric is utilized to measure the similarity between image patches, with lower values indicating greater similarity between two patches.
For dense pose estimation, the Mean Squared Error (MSE) is used to measure the distance between the ground truth and the predicted dense pose, with lower values indicating better quality.
For 2D keypoints, we adopt the Percentage of Correct Keypoints (PCK) which measures the percentage of predicted 2D keypoints that are located within a certain distance threshold from the ground truth. 
Specifically, we set the distance to $0.2 \times$ torso diameter (PCK@0.2).
For 3D human pose estimation, we use the Mean Per Joint Position Error (MPJPE), which calculates the mean Euclidean distance between the estimated and actual 3D joint positions.
Specifically, it is defined as:
\begin{equation}
\text{MPJPE} = \frac{1}{N} \sum_{i=1}^{N} \|\mathbf{J}_i^{\text{pred}} - \mathbf{J}_i^{\text{gt}}\|_2,
\end{equation}
where $N$ is the total number of joints, $\mathbf{J}_i^{\text{pred}}$ is the predicted 3D coordinate of the $i$-th joint, $\mathbf{J}_i^{\text{gt}}$ is the ground truth 3D coordinate of the $i$-th joint, and $\|\cdot\|_2$ denotes the Euclidean distance (L2 norm).
The MPJPE value is obtained by computing the Euclidean distance between the predicted and actual joint positions for each joint and then determining the average distance across all joints. 
Lower MPJPE indicates better performance because it indicates that the predicted joint positions are closer to the ground truth.

\begin{figure*}[tb]
    \centering
    \includegraphics[width=1\linewidth]{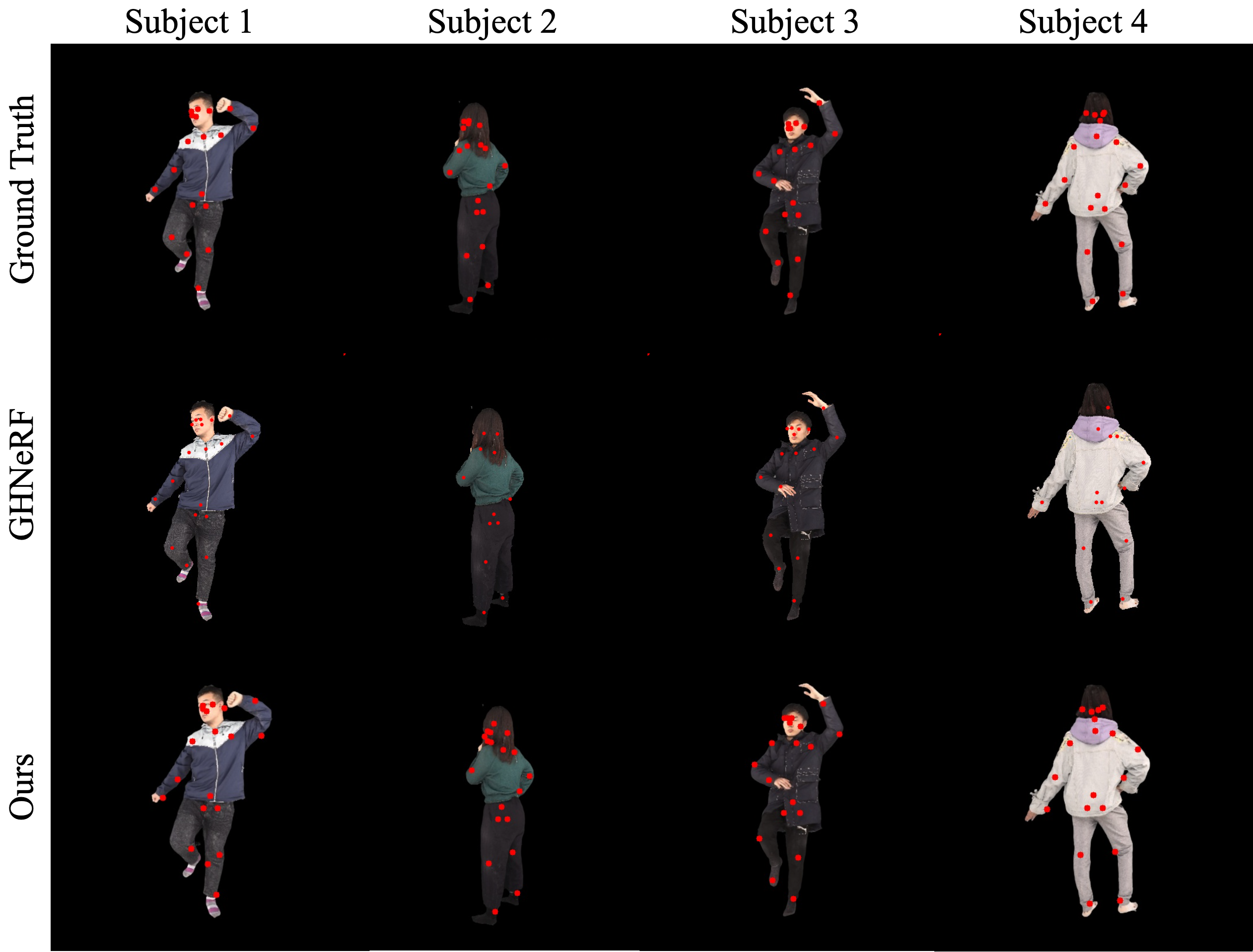}
    \caption{Comparison of qualitative results for the 2D keypoint estimation task on the THuman2.0 dataset.}
    \label{fig:keypoint_res}
\end{figure*}

\begin{figure*}[tb]
    \centering
    \includegraphics[width=1\linewidth]{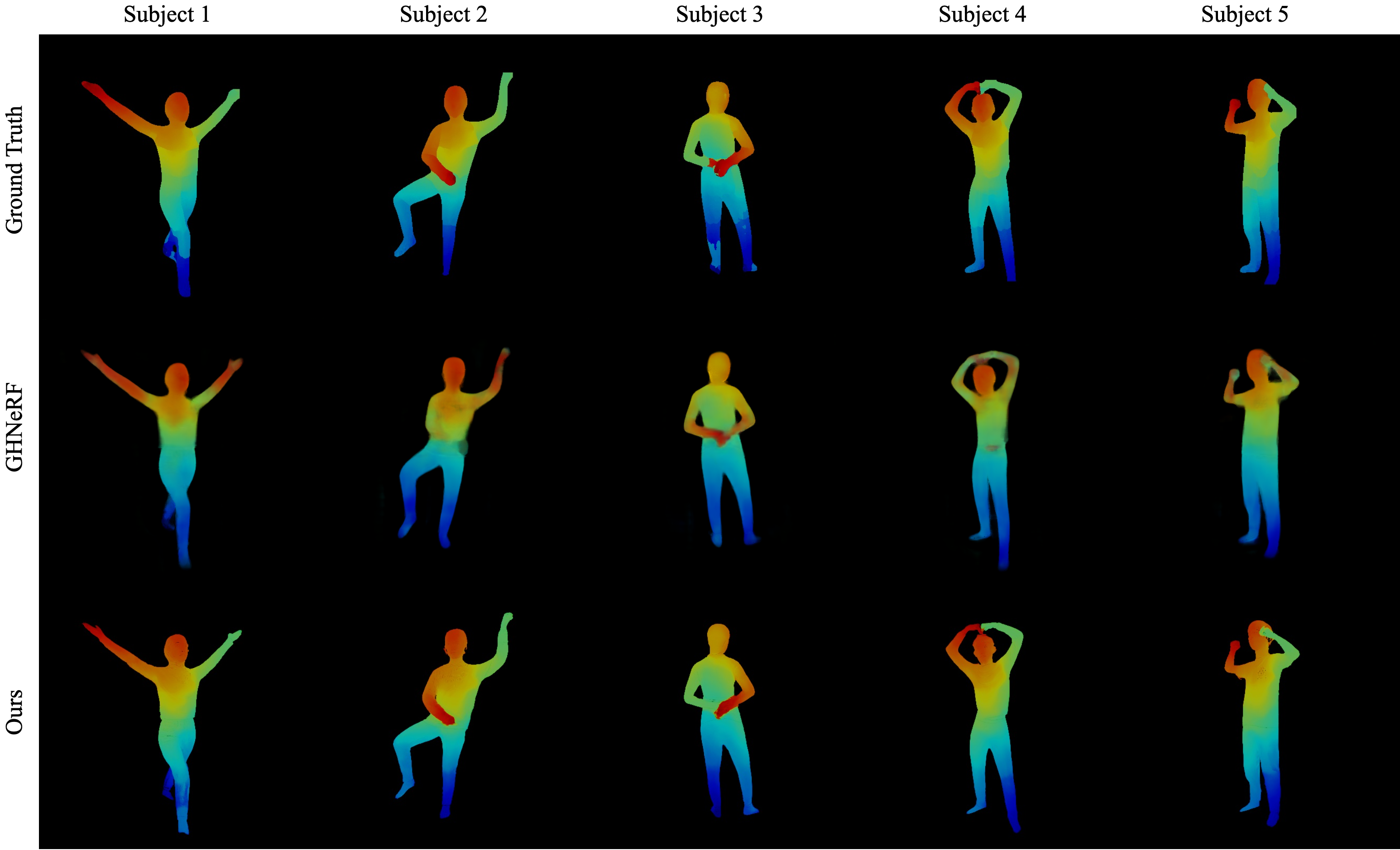}
    \caption{Visual comparison of dense pose estimation results. The findings indicate that our proposed method considerably outperforms GHNeRF~\protect\cite{Dey_2024_CVPR} in dense pose estimation.}
    \label{fig:dense_res}
\end{figure*}

\begin{figure*}[tb]
    \centering
    \includegraphics[width=1\linewidth]{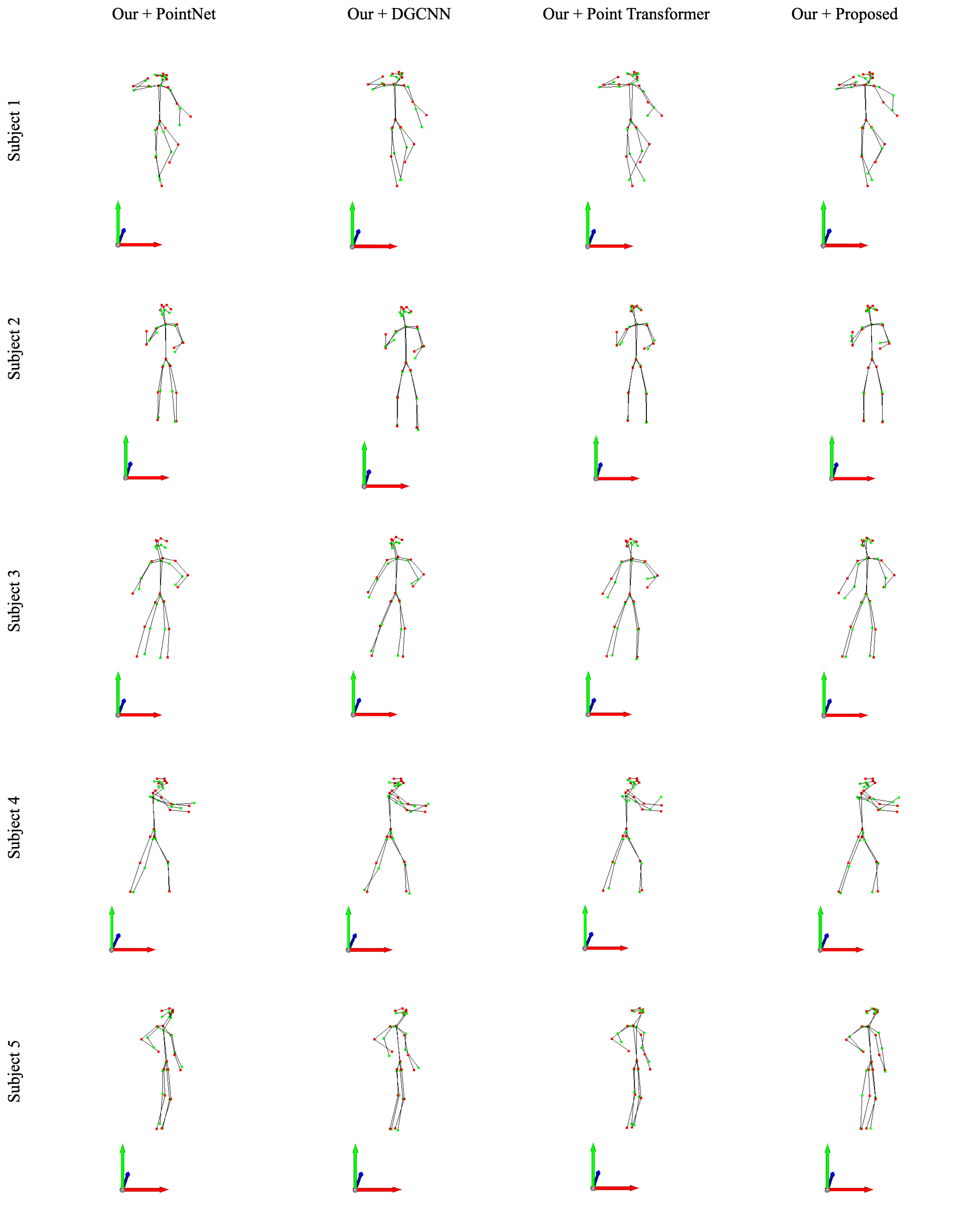}
    \caption{Qualitative results of 3D pose estimation using various pose regression network. The red markers show the actual 3D poses, while the green markers show the predicted 3D poses.}
    \label{fig:3d_pose}
\end{figure*}

\section{Limitations}

Although the current method provides substantial advancements compared to the state-of-the-art, it is limited to single-human representations and is primarily applicable to human subjects. 
Future research can explore extending this method to handle multi-subject representations and developing more generalized models applicable to a broader range of articulated objects or animals.

\begin{figure*}[tb]
    \centering
    \includegraphics[width=1\linewidth]{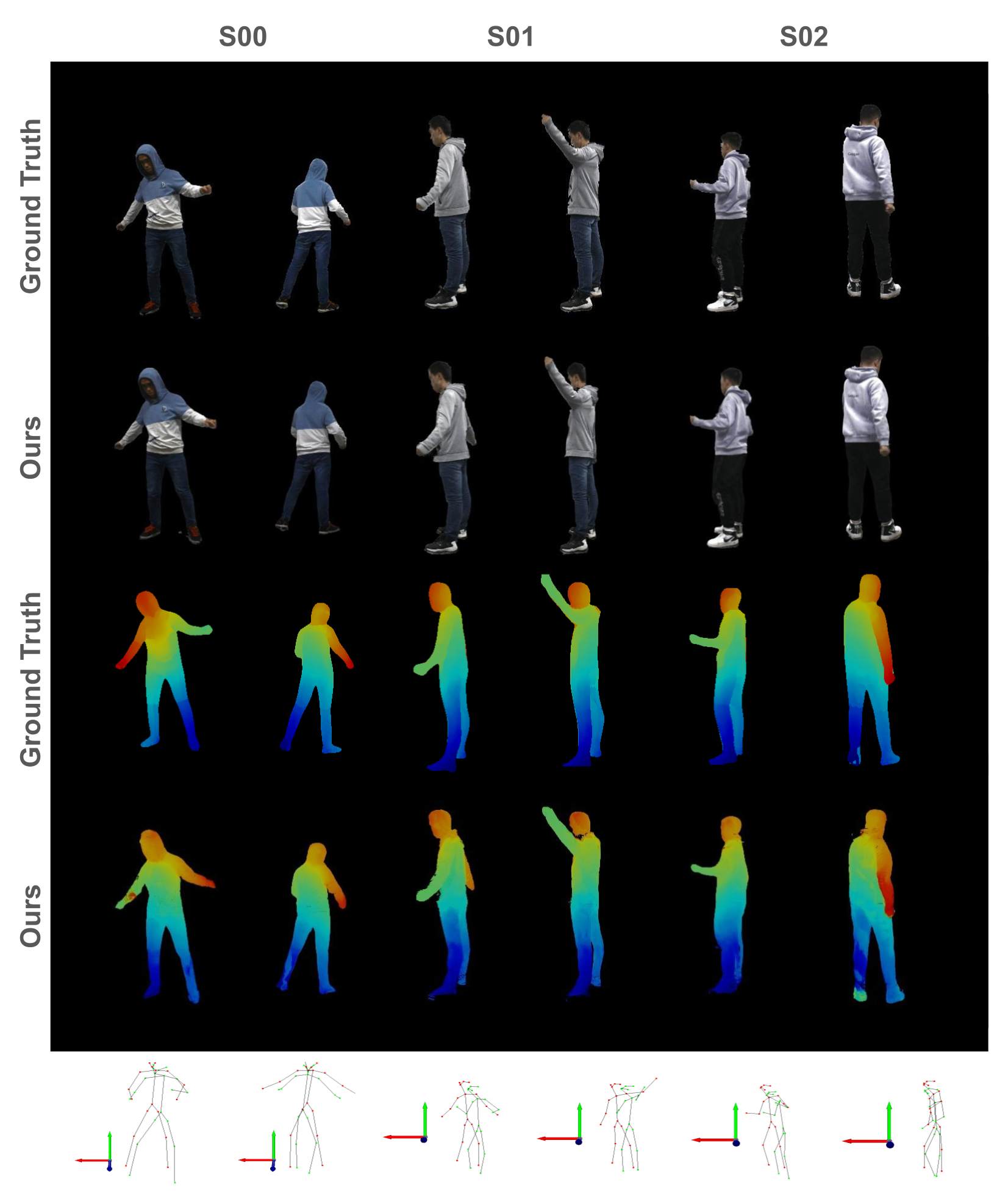}
    \caption{Qualitative results of HFGaussian on THuman4.0 dataset. For 3D pose estimation, the red markers show the actual 3D poses, while the green markers show the predicted 3D poses.}
    \label{fig:th4_res}
\end{figure*}

\begin{figure*}[tb]
    \centering
    \includegraphics[width=0.8\linewidth]{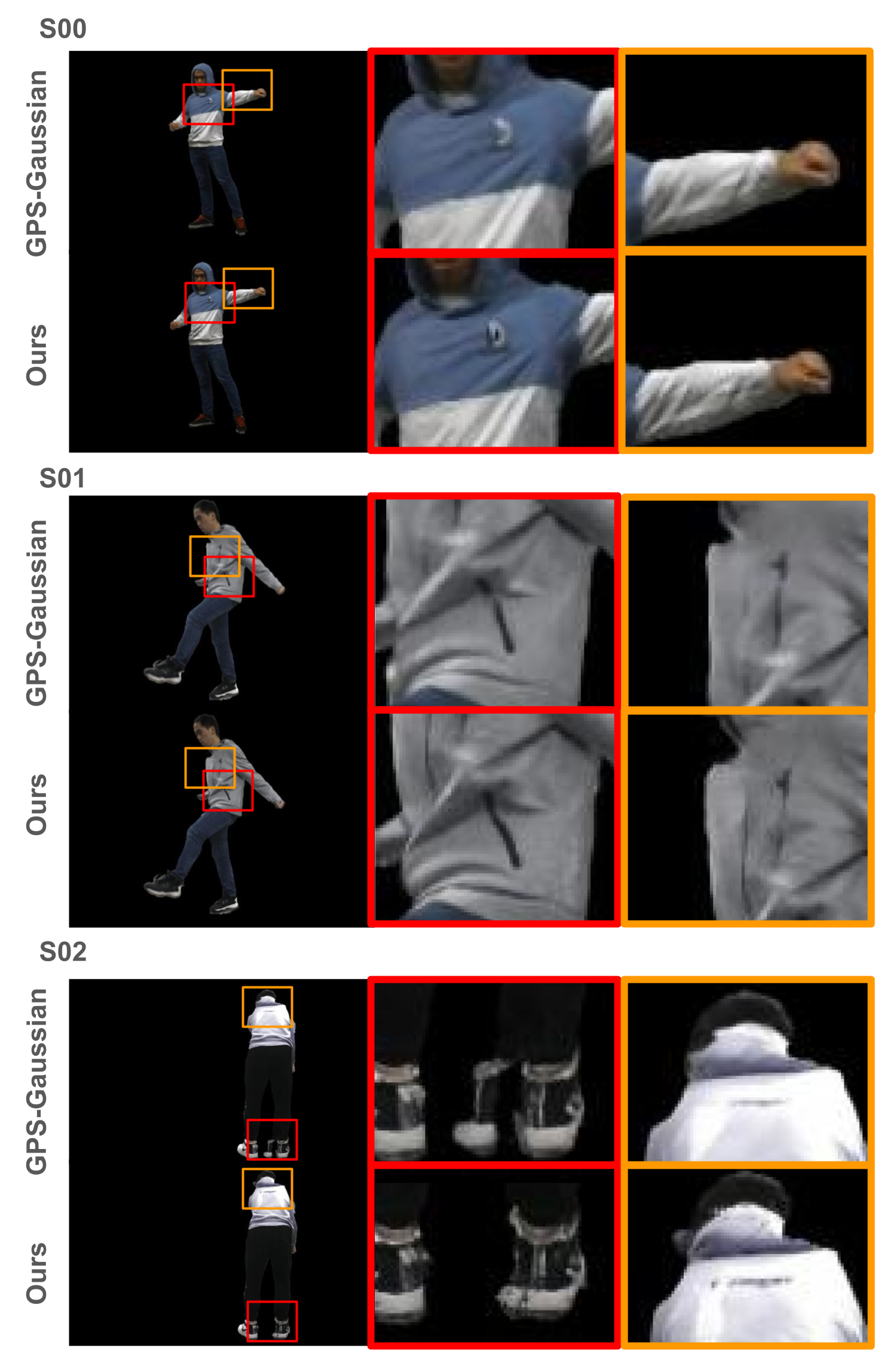}
    \caption{Qualitative results of HFGaussian and GPS-Gaussian on THuman4.0 dataset. GPS-Gaussian produces over-smoothed rendering results on the THuman4.0 dataset.}
    \label{fig:th4_res_cmp}
\end{figure*}

\section{Additional Results}
\label{add_res}

Figure~\ref{fig:keypoint_res} shows the 2D keypoint results comparing GHNeRF, the only existing method capable of rendering both RGB and human features simultaneously, with HFGaussian. 
HFGaussian is competitive and even outperforms GHNeRF in certain cases (e.g., Subject 4), while still maintaining real-time rendering speed.
For dense pose results, HFGaussian clearly outperforms GHNeRF in Figure~\ref{fig:dense_res}. 
The left hand of Subject 1, both hands of Subject 2, the right hand of Subject 3, and the hands and waist of Subject 4 are much closer to the ground truth. 
This level of detail in the hands highlights the superior capability of HFGaussian in accurately capturing fine human features.

Figure~\ref{fig:3d_pose} demonstrates that the proposed pose regression network can maintain good quality and rendering speed (see Figure 5).
This further verifies the statement that running the pose regression network on a subset of the 3D Gaussian is sufficient.

Table~\ref{tab:compare_method_th4} demonstrates that HFGaussian is more robust to unseen data. 
It outperforms all baselines by at least 0.5 PSNR across all subjects on RGB images, and shows a clear advantage over GHNeRF in capturing human features. 
Figure~\ref{fig:th4_res} and Figure~\ref{fig:th4_res_cmp} further highlight HFGaussian's robust generalization ability across all three subjects. 
Note that the ground truth 3D pose does not perfectly align with the predicted 3D pose due to the real-world camera poses not forming a perfect circle, which may cause slight shifts between them. 
However, the 3D poses are still relatively well-matched.

\end{document}